\documentclass[10pt,twocolumn,letterpaper]{article}

\usepackage[pagenumbers]{cvpr} % To force page numbers, e.g. for an arXiv version

\usepackage{graphicx}
\usepackage{amsmath}
\usepackage{amssymb}
\usepackage{booktabs}

\usepackage[pagebackref,breaklinks,colorlinks]{hyperref}

\usepackage[capitalize]{cleveref}
\crefname{section}{Sec.}{Secs.}
\Crefname{section}{Section}{Sections}
\Crefname{table}{Table}{Tables}
\crefname{table}{Tab.}{Tabs.}

\def\cvprPaperID{2125} % *** Enter the CVPR Paper ID here
\def\confName{CVPR}
\def\confYear{2023}

\begin{document}

\title{Neural Cell Video Synthesis via Optical-Flow Diffusion}

%%\and % TODO: add more people
%%Second Author\\
%%Institution2\\
%%First line of institution2 address\\
%%{\tt\small secondauthor@i2.org}

% \author{Manuel Serna-Aguilera\\
% University of Arkansas\\
% 1 University of Arkansas\\
% {\tt\small mserna@uark.edu}
% \and
% Khoa Luu\\
% University of Arkansas\\
% 1 University of Arkansas\\
% {\tt\small khoaluu@uark.edu}
% \and
% Nathaniel Harris\\
% University of Arkansas\\
% 1 University of Arkansas\\
% {\tt\small nqharris@uark.edu}
% \and
% Min Zou\\
% University of Arkansas\\
% 1 University of Arkansas\\
% {\tt\small mzou@uark.edu}
% }

% more compact
\author{Manuel Serna-Aguilera$^{1}$, Khoa Luu$^{1}$, Nathaniel Harris$^{2}$, Min Zou$^{2}$\\
$^{1}$CVIU Lab, University of Arkansas \\
$^{2}$Dep. of Mechanical Engineering, University of Arkansas \\
\tt\small \{mserna, khoaluu, nqharris, mzou\}@uark.edu,  
\vspace{-2mm}
}

\maketitle

%%%%%%%%% ABSTRACT
\begin{abstract}
The biomedical imaging world is notorious for working with small amounts of data, frustrating state-of-the-art efforts in the computer vision and deep learning worlds. With large datasets, it is easier to make progress we have seen from the natural image distribution. It is the same with microscopy videos of neuron cells moving in a culture. This problem presents several challenges as it can be difficult to grow and maintain the culture for days, and it is expensive to acquire the materials and equipment. In this work, we explore how to alleviate this data scarcity problem by synthesizing the videos. We, therefore, take the recent work of the video diffusion model to synthesize videos of cells from our training dataset. We then analyze the model's strengths and consistent shortcomings to guide us on improving video generation to be as high-quality as possible. To improve on such a task, we propose modifying the denoising function and adding motion information (dense optical flow) so that the model has more context regarding how video frames transition over time and how each pixel changes over time.

\end{abstract}

%%%%%%%%% BODY TEXT
%========================================================================
% 1. Introduction
%========================================================================
\section{Introduction}
\label{sec:intro}

Many problems and solutions in the computer vision and deep learning communities are geared toward natural images or video. A key component to solving these problems is to have as much data as possible. A problem domain very notorious for not allowing abundant access to data is biomedical imaging. We take the recent work of the video diffusion model presented by Ho et al. \cite{ho2022video} away from the natural image and video distributions and instead explore how to cater this model to biomedical images. For this work, in particular, we look to phase contrast microscopy videos of neural stem cells and strive towards generating high-quality \textit{synthesized} video data from Gaussian noise, effectively surpassing the data quantity problem. While this work achieves great results, for the domain of neuron cell videos, however, the current work misses some critical events and does not synthesize important structures well (as we will see later). We attempt to correct those mistakes in this work.

% describe problem of *creating* cell data
In the field of biomedical imaging, there are many reasons data collection can be a challenging task. In our case, growing and observing cells can be costly and slow. The timeline starting from proliferation (i.e., cell ``birth'') to cell maturity, and eventually to cell death, is a time-consuming process that can last 4 to 6 weeks for mature neuron cells. There is the potential for several issues to arise. Among those is the issue of phototoxicity, which may kill cells if exposed to high-intensity light and image sampling too frequently. A researcher must be consistently present to monitor the cell culture and ensure it is well-sustained and healthy every day that it is alive. Cell material also costs money. These problems severely limit how much data can be collected, and the lack of plentiful data limits how the deep learning community can tackle this problem domain.

% describe biomedical image data
Indeed, publicly available datasets for cells are limited for deep learning. Among these relatively-limited datasets is LIVECell \cite{livecell2021}, a dataset of approximately 1.6 million cells across 5,239 phase contrast images. Note that this dataset is recent (from 2021) and is, to our knowledge, the most comprehensive dataset of phase contrast images for detection and segmentation. Other ``predecessor'' datasets include EVICAN \cite{evican2020}, a dataset that is smaller in scale than LIVECell with 4,600 images with approximately 26,000 individual cells, and CellPose \cite{cellpose2020}, a relatively small public dataset of several types of microscopy images. 

% describe biomedical video data
For videos, the number of datasets featuring cells is also relatively few. A dataset from the biomedical community, called the ISBI Cell Tracking Challenge \cite{cell_track_challenge_2017}, is a collection of videos of several lines of cells (different types of cells) donated from several laboratories around the world. This data contains both 2D and 3D videos. For the 2D videos (relevant to our problem), there are 52 videos, where each cell line has four videos. Another prominent cell video dataset is the CTMC \cite{ctmc_2020} dataset, from 2020, made to address the shortage of datasets that facilitate analysis of cells moving in a culture. CTMC is comprised of 86 videos of cells from 14 different lines. (Note that these videos' frames are differential interference contrast--DIC--microscopy, which is different from phase contrast microscopy).

% compared with "natural image/video" datasets
The aforementioned datasets contrast with those describing the natural image world, where data collection is comparatively easy and relatively inexpensive (e.g., taking photographs of scenes or everyday objects, recording people in specific scenarios, etc.). It is what enables huge datasets such as COCO (2014) \cite{coco2014}, with approximately 328,000 images with 2.5 million instances of objects, MOT \cite{mot15, mot20}, etc., to exist. Meanwhile, public datasets of cells are considerably less numerous and with relatively fewer images, which impedes novel deep-learning progress in the biomedical imaging field. Factors like these make it difficult to collect large samples to create a dataset comparable to real-world images or videos.

% overview of the solution and what we do in this paper
\noindent \textbf{Contributions of this Work:} In this work, we propose to use the recent work of video diffusion and adapt it to synthesize neurons and support cells where: (i) the cells look realistic, (ii) cells in the culture move about the culture realistically, and, most importantly, (iii) cell-cell interaction has appropriately been learned. To achieve all three goals, we first consider incorporating dense motion (optical flow) into the denoising neural network to learn movement from the frames and use the pixel-level flow to learn the motion distribution. The idea is that the video and motion can be learned separately, and then we combine this knowledge to synthesize high-quality data. We also highlight the strengths and shortcomings of the baseline and modified approaches and offer advice for future synthetic dataset generation for biomedical images.

%========================================================================
% 2. Related Work
%========================================================================
\section{Background}

\begin{figure*}
    \centering
    \includegraphics[scale=0.45]{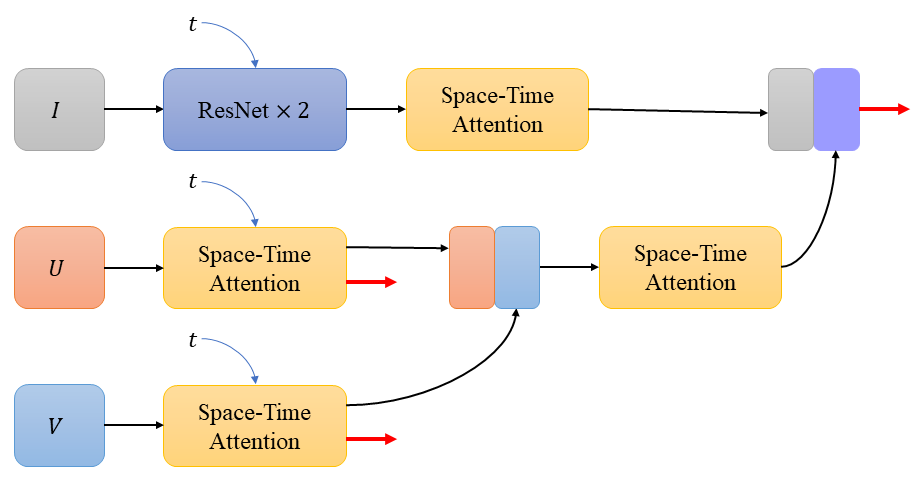}
    \caption{
        The proposed architecture for each scale block in the encoder, middle block, and decoder inside the denoising 3D U-Net within video diffusion. Our inputs are the video $I$ and the dense flow $U, V$ tensors of some spatial size $N \times N$ with $C$ channels. Each tensor is processed in its own branch, incorporating the time embedding $t$ in the first ResNet layers and space-time attention blocks (for the motion information). The dense information is merged by channel-wise concatenation, processed again, and concatenated with the video information. The bold arrows denote the output at the scale block, with the spatial dimensions doubled or halved, and the number of channels increased or decreased. When we pass the output of the \textit{entire} model, we get slightly denoised versions of our input data back.
    }
    \label{fig:proposed_method_overview}
\end{figure*}

% background--cell stuff
\subsection*{Materials and Cell Culture}
The primary subjects of our videos are neuron cells and several types of support cells, generally called glia (which we will henceforth refer to simply as ``support cells''). These cells are grown from stem cells called hippocampal human neural stem cells (hNSC's) from PhoenixSongs Biologicals, a biotechnology company. These cells are grown and kept alive in a controlled environment called a \textit{cell culture}; the cells are given supplements that keep them healthy and alive. Neurons are of interest due to their unique morphology and function. Neurons carry electrochemical signals in the brain by projecting neurites between cells to facilitate intercellular communication (note that neurites are part of the cell bodies). These extensions are 1-2 micrometers in width, can span distances of several hundred micrometers, and enable the cells to translocate and create a communication network with neighboring cells.

Once the cells are seeded down in the culture, they can be imaged \textit{in-situ}, or ``on-site'', with an Etaluma LS560 small form factor microscope capable of fitting inside the HERAcell Vios 160i incubator (the device to house the cell culture). Pre-centered phase contrast at 10$\times$ magnification enabled time-lapse observations of cells during culture. Timelapse sequences were captured with minimal brightness (9-15\% of a 1-watt LED light course) and long exposure times (200 ms), balanced with a high-frequency capture rate every 2-10 minutes to help mitigate phototoxicity. A work analyzing hippocampal neural stem cells \cite{cell_culture_setup2014} further describes the same cell culture setup we used for this work.

\subsection*{Diffusion Models}
% overview--since only some people are familiar with what exactly diffusion in this context is!
Diffusion models, first formulated several years ago \cite{thermodynamics_2015}, have recently risen into prominence after several framework improvements \cite{openai2021diffusion, ho2020denoising}. Nowadays, these models can synthesize better samples from a training distribution compared to GANs \cite{gan_2014} despite being slower and also tricky to control. It is also worth noting that diffusion models are used for more than just generating images. Other recent works train on different distributions like motion sequences \cite{motion_diff2022}, or even chemical-level interactions \cite{diffdock_2022}.

% forward noising process
To learn how to synthesize synthetic images, diffusion models do so by repeatedly and gradually adding Gaussian noise to the input data in a \textit{forward} process $q$. Formally, using the notation for the distributions from \cite{ho2022video},

\begin{equation}
    q(\mathbf{x}_t | \mathbf{x}_{t-1}) := \mathcal{N}\big(\mathbf{x}_{t-1}; (\alpha_t / \alpha_{t-1})\mathbf{x}_{t-1}, \sigma^2_{t | t-1} \mathbf{I} \big)
    \label{diff_forward_process}
\end{equation}

\noindent
where $t$ is the current iteration, $\mathbf{x}_t$ is the latent at iteration $t$ (we can let $\mathbf{x}_t$ where $t=0$ denote the input data) up to iteration $T$, and $\alpha_t$, $\sigma_t$ give the variance schedule for iteration $t$. The variance scalar $\sigma^2_{t | t-1} = (1 - e^{\lambda_t - \lambda_{t-1}})\sigma^2_{t}$ where $\lambda_t = \log(\alpha^2_t / \sigma^2_t)$ is the log signal-to-noise ratio which decreases with increasing $t$. The scalars can be manually defined. Note that, unlike VAEs and GANs, the latent representations are the same dimensionality as input data $\mathbf{x}_0$. Note $\mathbf{x}_0 \sim p(\mathbf{x})$, i.e., the input image belongs to a particular distribution (in our case, the distribution of phase contrast images with neurons), and the last latent is approximately Gaussian at iteration $T$, i.e., $\mathbf{z}_T \sim \mathcal{N}(\mathbf{z}_T; \mathbf{0}, \mathbf{I})$. 

% backward/denoising process
The backward process, denoted by $p$, takes in Gaussian noise of the same dimensionality as $\mathbf{x}_0$ and gradually removes the noise to construct a sample from the learned training distribution, i.e., we get a sample $\mathbf{x}_0 \sim p(\mathbf{x})$ by only giving Gaussian noise. Formally, to predict the next latent given the previous latent, using the notation of \cite{ho2020denoising},

\begin{equation}
    p_{ \theta }(\mathbf{x}_{t-1} | \mathbf{x}_t, \mathbf{x}_0) := \mathcal{N}\big(\mathbf{x}_{t-1}; \tilde{\mu}_{t-1 | t}(\mathbf{x}_t, \mathbf{x}_0), \tilde{\sigma}^2_{t-1 | t} \mathbf{I}
    \big)
    \label{diff_backward_process}
\end{equation}

\noindent
It tells us that we can learn what noise to remove at each step given Gaussian noise at iteration $t=T$ until we get an approximation of a sample from the training distribution. A neural network learns this denoising process with parameters $\theta$ to get an estimate for $\mathbf{x}_0 \sim p(\mathbf{x})$.

\subsection*{Video Diffusion Model}
% overview of video diffusion
The video diffusion model, as described by Ho et al. \cite{ho2022video}, naturally extends the idea of synthesizing images towards synthesizing videos. Instead of taking in a batch of training images, it processes a batch of videos with the shape (frames $\times$ height $\times$ width $\times$ channels), where each video has a constant number of frames. As in \cite{ho2020denoising}, the neural network used to learn denoising for reconstructing images is a modified 3D U-Net. The data is passed through a 2D convolutional layer in each block, and then a spatial attention operation is performed. For videos, 2D convolutions become 3D (1 $\times$ 3 $\times$ 3), and a temporal attention operation is added as the last operation, which operates over the dimension of the frame. To train this particular diffusion model for generating videos, Ho et al. \cite{ho2022video} use a weighted squared error (L2) loss, taking the difference between the input video and the synthesized video. We could also use a regular L1 loss.

%========================================================================
% 3. New Method: Enforcing Realistic Cell Movement
%========================================================================
\section{The Proposed Method}

% what model are we using?
We now describe our modified approach for video diffusion that enforces dense motion to synthesize more realistic cell movement and cell-cell interaction. We draw upon several works \cite{fusionseg_2017, motion_unet2020, motion_diff2022} and adapt their design philosophies to modify the denoising model to handle motion and video at the same time.

% overview of method--what I modified--without too many details
Within the modified 3D U-Net, we reduce the number of channels of the image the model processes from 3 to 1. Phase contrast microscopy images are grayscale, and we fill in that free space with motion information. To accommodate this new data more appropriately, we create independent processing branches for the image and optical flow information. The following sections describe the different components of the new approach.

\subsection{Introducing Motion Information} % how do we get optical flow?
We thought to first relieve the computational burden of the denoising U-Net from learning the cells' motion from scratch so it could focus on generating video. To do this, we introduce dense optical flow information to each frame in a given video. Optical flow, i.e., computing the velocities of each pixel for a given pair of consecutive frames, is still a researched topic in computer vision. To simplify matters, we use Horn-Schunck (HS) optical flow \cite{hs_flow1981} to derive our motion information in the $x$ and $y$ directions. Given an input video of cells moving, and for each consecutive pair of frames $k$ and $k+1$, we compute the $x$-velocities $U^{(k)} \in \mathbb{R}^{N \times N}$ and the $y$-velocities $V^{(k)} \in \mathbb{R}^{N \times N}$, where $N$ is the height and width of the videos (we will be working with square frames in all formulations and experiments). Via the HS method, we compute the matrices for flow for frame $k$ below.

\begin{align}
    U^{(k)} &= \Bar{U}^{(k)} - \frac{I^{(k)}_x \odot \Bar{U}^{(k)} + I^{(k)}_y \odot \Bar{V}^{(k)} + I^{(k)}_t}{\lambda^{-1} + (I^{(k)}_x)^2 + (I^{(k)}_y)^2} \odot I^{(k)}_x \label{hs_u} \\
    V^{(k)} &= \Bar{V}^{(k)} - \frac{I^{(k)}_x \odot \Bar{U}^{(k)} + I^{(k)}_y \odot \Bar{V}^{(k)} + I^{(k)}_t}{\lambda^{-1} + (I^{(k)}_x)^2 + (I^{(k)}_y)^2} \odot I^{(k)}_y \label{hs_v}
\end{align}

\noindent
Note that the symbol $\odot$ is for element-wise multiplication, and a squared matrix is multiplied element-wise with itself. $\Bar{U}$ and $\Bar{V}$ denote the matrices of the local average for each pixel, defined for each pixel $(i,j)$ as $\frac{1}{4}\big(U^{(k)}(i+1,j) + U^{(k)}(i-1,j) + U^{(k)}(i,j+1) + U^{(k)}(i,j-1) \big)$ and similarly for $\Bar{V}$; initially, $\Bar{U}^{(k)}=\mathbf{0}, \Bar{V}^{(k)}=\mathbf{0}$. In our computations, the image gradients $I_x$ and $I_y$ are derived by performing convolution with a Sobel-$x$ $(3 \times 3)$ filter and Sobel-$y$ $(3 \times 3)$ filter, respectively. The temporal derivative $I^{(k)}_t$ is computed via the forward difference (i.e., take the current frame and subtract the next frame from it). The term $\lambda$ is the weight term for the brightness constancy assumption for HS flow. We then take, for each video, all the $U^{(k)}$'s and $V^{(k)}$'s to form the video motion tensors $U, V \in \mathbb{R}^{K \times N \times N \times 1}$, where $K$ denotes the total number of frames in the input video sample, and we explicitly state it is single-channel.

Introducing this data effectively means that the video diffusion model not only learns the image distribution of phase contrast microscopy but the \textit{dense motion distribution} as well. Thus, the model can learn to synthesize correct frames and learn how these cells move (or, more formally, learn the motion distribution from the perspective of optical flow).

\subsection{Video and Motion Diffusion Model}
% input and design philosophy/justification
Now we describe the modifications to the 3D U-Net in video diffusion. For our approach, each tensor per channel carries different information, similar to \cite{motion_unet2020} to help the model learn from the extra information, not just from the video. The input for our modified denoising model is of shape $\mathbb{R}^{K \times N \times N \times 3}$. In the first channel index, we have the grayscale video data (pixel values are all equal across the RGB channels in the original videos). The following two channels provide dense optical flow information. In the second channel index, we have the dense optical flow data in the $x$-direction for the video, $U$, and the third channel index contains the dense optical flow data for the $y$-direction for the video, $V$.

% image branch
Given that we have three different pieces of information, we follow the general design philosophy of \cite{fusionseg_2017} to process each piece of data \textit{independently} before combining data at each scale level. The idea of processing each piece of information separately means the denoising U-Net can learn cellular motion patterns in both $x$ and $y$ axes more appropriately and then incorporate that knowledge into the video data. To that end, we process the video information in its branch; this branch takes in input $I \in \mathbb{R}^{K \times N \times N \times 1}$. This video data is treated in the same manner as in the original denoising U-Net, where we pass it with the time embedding at the denoising iteration $t$. Initially, we treat the video data with convolution and initial temporal attention layers before passing it off to the first block. The video data is passed through two ResNet layers at each scale level: spatial and temporal. It is the same in the encoder, middle block, and decoder portions; the only difference is that the data is single-channel. We still need to downsample or upsample, as we must also process the motion data at the same scale level first.

% flow branches
Now that we also pass the motion information separately, we have the model independently learn the dense motion distribution. We adopt a similar design in \cite{motion_diff2022} and process the motion mostly with factored space and time attention layers. Learning features from the motion data with ResNet layers does not seem reasonable to perform on non-image data. We only use single convolutional layers to increase or decrease the number of channels in these branches. After processing, the dense directional motion is passed on to the following scale block, where all data is spatially downscaled. The directional dense motion branches are also merged via channel-wise concatenation. This merged data is passed through another space-time attention block, and the number of channels is halved. 

% merging and overall processing of data
After processing the video and flow data in their branches at each scale level, we take both branches' output and merge them via channel-wise concatenation. This resulting tensor is assigned back into the video information $I$, and the last operation for the video branch is halving the number of channels with a convolutional layer. Finally, the outputs are downsampled via 3D convolution, like in the original work. The downsampled or upsampled data is then passed on to the next scale-level block in the U-Net.

%========================================================================
% 4. Experiments & Results/Discussion
%========================================================================
\section{Experimental Results}
\subsection{Raw Video Data}

\begin{figure}[t]
    \centering
    \includegraphics[width=0.8\linewidth]{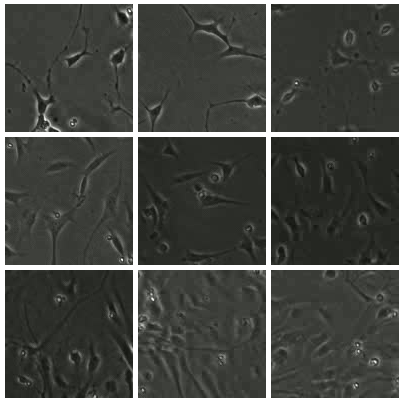}
    \caption{
        A visualization of the $(128 \times 128)$ training video dataset. We emphasize variety with the cell culture density, cell maturity, and mix of neurons and support cells.
    }
    \label{fig:data_vis}
\end{figure}

\begin{figure*}
    \centering
    \includegraphics[width=1.0\linewidth]{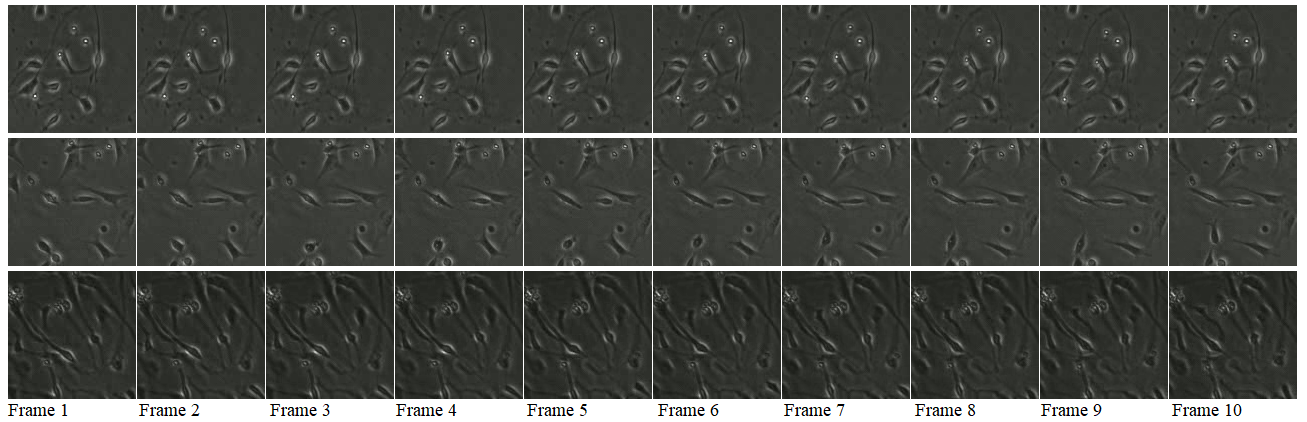}
    \caption{
        Three real video samples, ten frames each. The first row is a video showing a less dense culture where the cells in the center move, changing shape as they go. The second row is a video showing cells in the center and bottom using their appendages to move. It is the cell-cell interaction we want to learn. The third row shows a denser culture where cells are huddled close together, and more mature cells do not move as drastically as in the other videos due to limited space in the substrate.
    }
    \label{fig:train_data_video}
\end{figure*}

% raw video data
To train our modified video diffusion model for cell movement, we have imaged ten cell cultures containing neuron cells, the various support cells, and the occasional mass of dead cell matter for several hours. Each video sample contains different cell culture densities and imaging timeframe for increased variability. For more information, \cref{tab:video_info} gives more details about each video; more information about comparing distributions can be found in the supplementary materials.

\begin{table*}
    \centering
    \begin{tabular}{lrrrrr}
        \hline
        \multicolumn{1}{c}{\textbf{Culture}} & \multicolumn{1}{c}{\textbf{Frames}} & \multicolumn{1}{c}{\textbf{Capture Time (hrs)}} & \multicolumn{1}{c}{\textbf{Day Imaged}} & \multicolumn{1}{c}{\textbf{Culture Density}} & \multicolumn{1}{c}{\textbf{Cell Maturity}} \\ \hline
            1 & 1728 & 4.5 & 5 & Moderate & Mature \\ \hline
            2 & 298 & 2.5 & 5 & Moderate & Mature \\ \hline
            3 & 446 & 18 & 0,1 & Moderate & Very Young \\ \hline
            4 & 1448 & 27 & 1,2,3 & Dense & Mature \\ \hline
            5 & 811 & 48 & 4,5 & Dense & Very Mature \\ \hline
            6 & 550 & 18 & 6 & Very Dense & Very Mature \\ \hline
            7 & 541 & 24 & 0 & Very Sparse & Very Young \\ \hline
            8 & 1448 & 48 & 1,2 & Very Sparse & Mature \\ \hline
            9 & 1448 & 48 & 4,6 & Very Sparse & Very Mature \\ \hline
            10 & 724 & 24 & 7 & Moderate & Mature \\ \hline
    \end{tabular}
    \caption{ Details on the different cell cultures grown (indexed by a number) where the input patch videos were derived. All original videos are $1400 \times 1400$. All RGB channel values for each pixel are the same. The ``Day Imaged'' column refers to how many days the culture was allowed to mature before imaging occurred (day 0 is initial cell culture seeding); the days are different for data variability.
    }
    \label{tab:video_info}
\end{table*}

\subsection{Patch Video Training Data}

% why we chose dimension and number of frames
Given GPU memory limitations when taking in videos, we needed to crop each high-resolution video into small patch videos. We decided that $(128 \times 128)$ and 120 frames for each patch video were enough to capture cell movement and limit GPU memory consumption. If we chose a size of $(64 \times 64)$, there would be data quality issues for too many samples. For instance, in many videos, we would observe a much higher rate of cells coming in and out of frame, which does not allow the model to learn cell motion properly. The cell bodies and neurites may also be much larger than an area of $(64 \times 64)$, improperly causing them to be cut off. Since we also want to capture cell morphology adequately, a small patch size may impede the learning of cell behavior. We would also observe many $(64 \times 64)$ videos containing no cell material. Therefore, despite the patch size $(64 \times 64)$ offering many more frames to work with as in the experiments in \cite{ho2022video}, we work with $(128 \times 128)$ for all experiments.

% how we computed the patches
For all videos of all cell cultures, the patches have 25\% overlap with another neighboring patch, i.e., when moving in the $x$-direction, and similarly for the $y$-direction, the next top-left corner of the next patch will be in location $x = x + aw_p$, where $x=0$, initially, up to $W-w_p$, where $W$ are the width of the video frame, $w_p$ is the patch width, and $a \in [0.0, 1.0]$ is the percent overlap. To produce much more patch videos from one HD video, each of the HD videos was sampled at different time stamps such that cells were in sufficiently different positions and at different maturity levels. Of note, for some cultures, like 1, 2, 6, and 7, cells moved slowly, and therefore frames from the raw videos were skipped to produce more apparent movement. Doing this, we have 2,194 patch videos for our training data.

% implementing optical flow
To compute the new data with dense optical flow information, the directional velocities defined by equations \ref{hs_u} and \ref{hs_v} in theory are computed until their elements converge. In practice, however, we only iterate ten times to get $U$ and $V$. The brightness constancy term $\lambda$ is set to 1.

\subsection{Implementation Details}
%\subsubsection*{Training Settings}
We only work with ten frames when reading the patch videos due to memory limitations. Similarly, we use the first ten frames when reading video with optical flow data for memory reasons. We opted to have a couple more samples per batch than have a couple of samples with a few more frames so the models could learn different scenarios from the training video distribution.

At training time, we normalize each sample to be within the range $[0.0, 1.0]$. We specifically normalize per channel (so we do not mix the video and motion data statistics). We also apply two geometric augmentations: horizontal and vertical flipping of the frames, each occurring with a probability of 50\%.

We let the original video diffusion model train for three days and 12 hours, or 500 epochs. Since the modified denoising model is significant, we let it train for two additional days. Each model was set to have a learning rate of 0.0001, with an Adam optimizer with $\beta_1=0.9, \beta_2=0.999$. We used an L1 loss for all the models. For the codebase, we adapted an unofficial implementation in PyTorch \cite{pytorch_lib} for the video diffusion model and made it run on multiple GPUs. For the original model, we trained on two Quadro RTX 8000 GPUs (48 GB RAM) with two samples per GPU.

\subsection{Cell Movement in Training Data}
For comparison purposes, we describe how neurons and their support cells should move in cell culture (in the training data). In our phase-contrast videos of cells, neurons and support cells may be stationary, move slowly, or move quickly. Neurons, in particular, extend a sensory receptor, called a growth cone, at the end of their axon to feel their extra-cellular environment, which includes input from the substrate, neighboring cells, chemical gradients, etc. This sensing function is critical in the cell deciding migration direction. To move, neurons and support cells repeatedly stretch and form temporary focal adhesions with the substrate, whereby they ``drag'' themselves along the surface. This process leads to shape change and usually causes the cell bodies to become slender. When we view more dense cell cultures where cells can interact, the neurons' appendages often appear to cling together for a time. In our collection of phase contrast images, one may observe clumps or ``balls'' with a bright center. These are the remains of dead cells and should not move unless they are physically pushed around by other cells (either the body or its appendages). We should not see these clumps move alone from a video synthesis standpoint. What is very tricky, however, is that sometimes these clumps look very similar to \textit{some} neurons moving about the culture. They appear to have a bright center, and their appendages used for moving are narrow as the density and maturity of the culture increase. They do not move as much anymore as they form a web-like network of neurons clinging to each other.

\begin{figure}[t]
    \centering
    \includegraphics[width=1.0\linewidth]{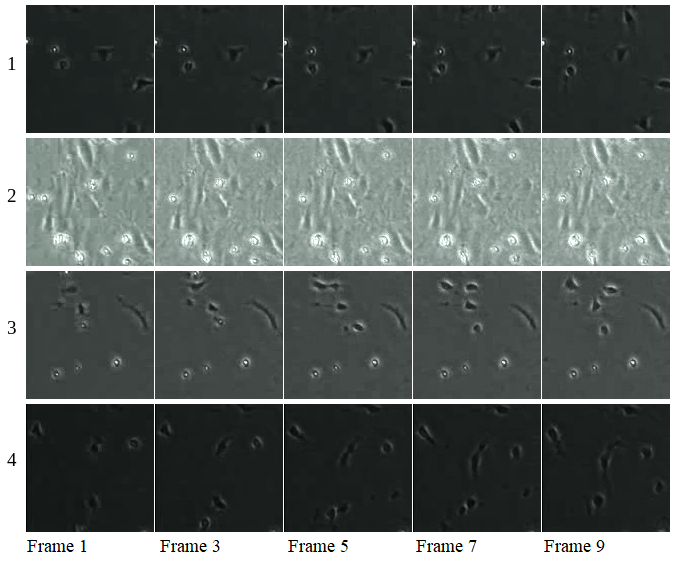}
    \caption{Typically synthesized videos from the baseline video diffusion model. Showing odd-numbered frames to conserve space. Row or video sequence \textbf{1} shows a good-quality video sequence of fake cells moving and their appendages being properly shown. Row \textbf{2} shows that the model struggles to synthesize denser cultures properly; the overall quality of these types of samples tended to be noisy. Row \textbf{3} shows a bad case. They are paying attention to the ``worm''-like cell towards the right. It may be a mature neuron or a neuron in the process of moving. However, the cell body shape and movement do not seem realistic. In Row \textbf{4}, notice the two ``cells'' at the bottom merging, with one simply disappearing. We want to avoid synthesizing these kinds of interactions.
    }
    \label{fig:baseline_results}
\end{figure}

\begin{figure}[t]
    \centering
    \includegraphics[width=1.0\linewidth]{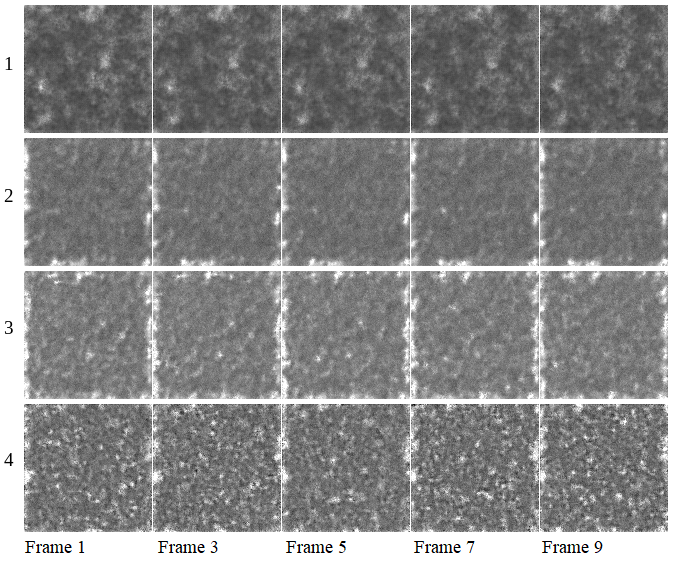}
    \caption{Synthesized results from mixing dense motion and video information. Showing odd-numbered frames to conserve space. As can be seen, the model gets close to generating a scene resembling a cell culture, but spots resembling cells appear.}
    \label{fig:proposed_method_samples}
\end{figure}

\subsection{Discussion: Baseline Synthesized Videos}

% cell area distribution comparison
\begin{figure}[t]
    \centering
    \includegraphics[width=1.0\linewidth]{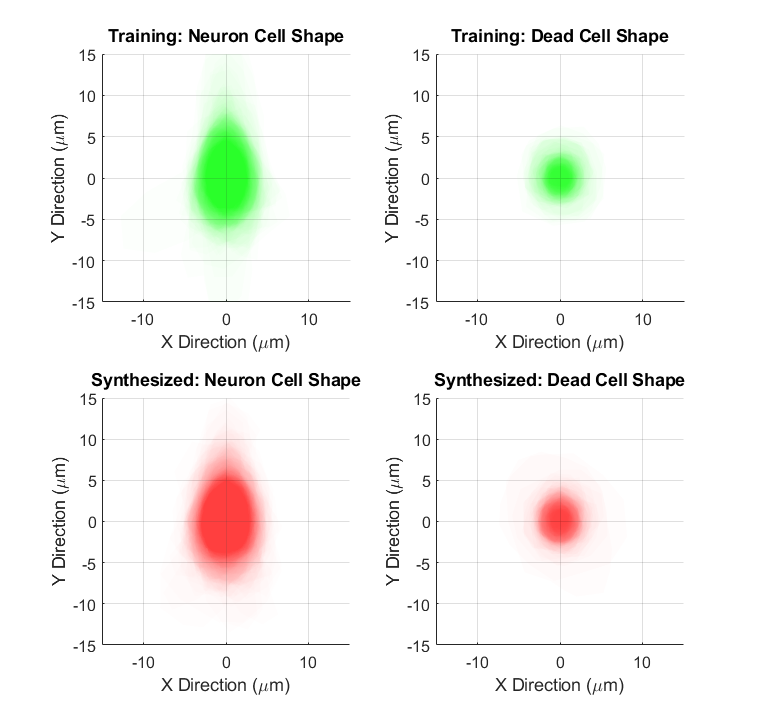}
    \caption{Distribution of cell and dead cell shapes across the synthesized results and the training patches. Shapes were annotated and overlaid to show the similarity of each set.}
    \label{fig:cell_area_comparison}
\end{figure}

% neurite direction
\begin{figure}[t]
    \centering
    \includegraphics[width=1.0\linewidth]{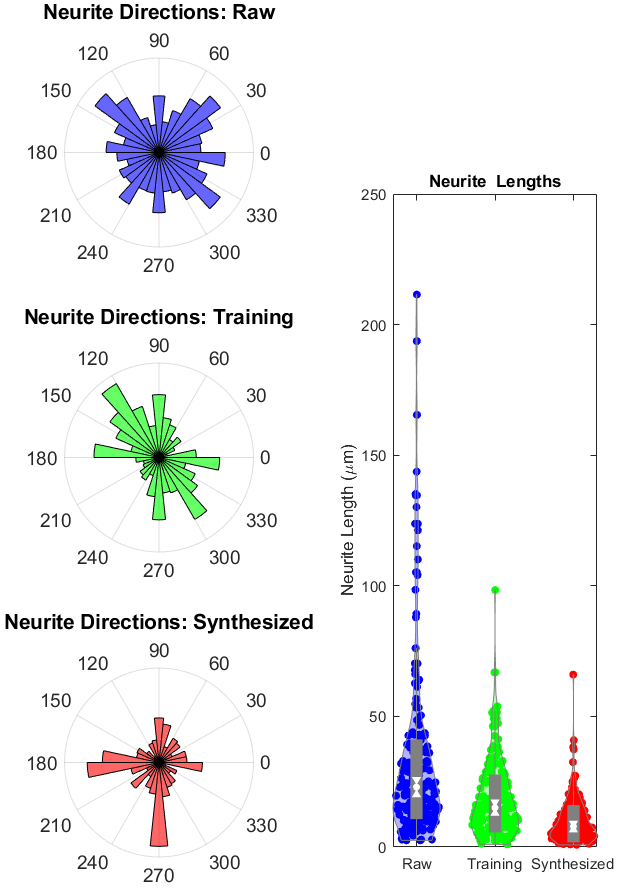}
    \caption{Distributions of neurite directions (left) and neurite lengths (right) comparing the raw data (blue), training data (green), and synthesized data (green). A grey boxplot in the center of the violin indicates the interquartile range, and the median and 95\% confidence intervals for each dataset are indicated with white circles and triangles, respectively. Clear differences in the distributions indicate little similarity and, therefore, room for improvement in the approach.}
    \label{fig:neurite_direction}
\end{figure}

% baseline discussion: sparse cultures
The synthesized results from training the video diffusion model qualitatively resemble real neurons in culture, but critical differences are somewhat subtle. The cells generally appear realistic for synthesized videos showing relatively few cells in the culture. They are small, exhibit the typical triangular or teardrop shape, and neurons reach out with their bodies to move between frames. In some cases, however, cells do not appear as they should. Some such cases show ``cells'' that are elongated for a long time. These abnormal objects may represent the model trying to create a cell with long neurites or a neurite disconnected from a cell body (for example, refer to the video sequence in row 3, figure \ref{fig:baseline_results}). Therefore, in our models, the synthesis process may confuse the different stages in a cell's life cycle. Several video samples correctly contain motionless chunks for samples containing small clumps of dead cell matter, while others show them moving on their own. It also does not mimic actual behavior (refer to the video sequence in row 3, figure \ref{fig:baseline_results}). In some video samples, some cells appear to divide, which seems to imitate a proliferation event in the culture, which is commonly observed in the early stages of the culture. We suspect the video diffusion model does not have enough frames to complete the action it was trying to synthesize, and therefore we cannot determine the result of these events. In other video samples, the opposite interaction happens, where cells appear to merge without any indication of the two cells still being present. This merging process is not natural, but it could be a case of the model being unable to complete the action of cells climbing over each other.

% baseline results: denser cultures
A significant proportion of the training patch videos contain dense cultures where the cells mature, and we expect video diffusion to synthesize such cases in approximately equal proportions. In our models, however, the synthesis of such cases yielded poor results. In our synthesized samples, we often generated fewer cells, with the background occupying a large area. Typically, for the synthesized dense cultures, individual cell morphologies are different from what they should be, and many samples contain what seems to be an excessive number of dead cell clumps. A typical example of dense culture generated from our models is in the second row in Figure \ref{fig:baseline_results}. 

% a more quantitative analysis
\begin{table}
    \centering
    \begin{tabular}{lrr}
        \hline
        \textbf{Distribution} & \multicolumn{1}{l}{\textbf{Neuron}} & \multicolumn{1}{l}{\textbf{Dead Cell}} \\ \hline
        Area                  & 0.105                               & 0.215                                  \\
        Perimeter             & 0.044                               & 0.077                                  \\
        Length                & $2.545   \times 10^{-10}$           & \multicolumn{1}{c}{--}                 \\ \hline
    \end{tabular}
    \caption{The $p$-values for comparing neurons and dead cells between the training patch data and the synthesized baseline videos based on different distributions. Note that the length distribution only applies to the living neurons.}
    \label{table:p-values-baseline}
\end{table}

% a more quantitative analysis
Quantitatively, video diffusion-synthesized cells and the cells from the video data are \textit{not} significantly different, statistically speaking. We performed two-sample t-tests for the cell areas, cell perimeters, and neurite length distributions (using the annotation procedure from Jung et al. \cite{cell_annotation1_2021}) between the synthetic videos and the training patch videos. Neuron cell bodies and neurites were traced using a GUI; their areas, perimeters, lengths, and directions the neurites were recorded. We used a 5\% significance level, with approximately 200 cells and 100 dead cells from the raw videos ($1400 \times 1400$), a sample of the training patch dataset, and the synthesized set of videos (both $128 \times 128$). The $p$-values for the measured area and perimeter (shown in Table \ref{table:p-values-baseline}) elucidate that the area distributions are highly similar to the training videos, whereas the perimeter distributions are less similar. In both cases, all distributions show $p > 0.04$. Comparing the shapes of synthesized and training datasets, shown in Figure \ref{fig:baseline_results} that the model accurately recreates the shapes and size of neurons and dead cells. For a more compact comparison of the shape distributions, Figure \ref{fig:cell_area_comparison} shows how well video diffusion has learned to synthesize shapes correctly. Details and visualization of the custom annotation can be found in the supplementary materials.

% major drawback--neurites!
An important limitation of the synthesized results is the direction of the neurites and their length. In the synthesized samples, neurites were short and skewed towards the four cardinal directions, whereas the raw dataset shows neurites extended in all directions, as shown in Figure \ref{fig:neurite_direction}. The training dataset also shows more skewness than the raw dataset, indicating the impact of the small frame size on properly capturing neurite direction and length. The neurites in the images are small features, and the resolution of the patches may be insufficient to estimate neurites of proper length. Some neurites in the raw data even span hundreds of pixels of distance. To compare the neurite length distributions, we also performed a two-sample t-test, where we computed a $p$-value $2.545 \times 10^{-10}$ against the training dataset, indicating the neurite length distributions are significantly different.
%It indicates the neurite length distributions are significantly different from what is desired, and further work is needed to address this critical aspect of cell culture data generation. 

\subsection{Discussion: Synthesized Videos from Modified Denoising Model}
% why the model does not work/limitations
After experimentation, we found that combining dense optical flow with learning to generate videos yielded a different result than we expected. As shown in Figure \ref{fig:proposed_method_samples}, most samples resemble a plain cell culture, but any ``cells'' that would be present are not discernable; all results were noisy. Therefore, we conclude that the U-Net needs to be overhauled or switched out with a more powerful denoising model to properly learn the motion between each frame in the training data, \textit{and} learn the variability for each scenario (of cell-cell interaction). Such a model would need to learn both the video and motion to produce realistic behavior and address the limitations of video diffusion.

% future work that will have to be done and ideas
For future work, we will have to abandon dense optical flow and focus on how features in each pair of consecutive frames change. Such a method that tracks features is the classic Lucas-Kanade approach, still prominent as its ideas are still incorporated in the deep learning community. This approach expects us not to rely on the motion data (which takes up memory) but instead learn how essential structures in each video change over time.

%========================================================================
% 5. Conclusions/Remarks on Cell Synthesis
%========================================================================
\section{Conclusion}
In this work, we have applied the recent video diffusion model to the biomedical field to phase contrast microscopy images of neurons. Video frames of the synthesized neurons were compared to actual images of neurons, and their size, area, perimeter, and neurite length and directions were compared. While the video diffusion model can synthesize high-quality videos of cells, some critical scenarios and behaviors must be consistently synthesized at high quality. Hence, we proposed modifications to the denoising 3D U-Net to handle spatial information and dense optical flow information, so the model has the context of action. Our hope with this analysis is to provide insight into how to cater this rising class of generative models to a problem domain infamous for data scarcity. 

%========================================================================
% Include references
%========================================================================
{\small
    \bibliographystyle{ieee_fullname}
    \bibliography{egbib}
}

\end{document}

\title{Supplementary Material for Neural Cell Video Synthesis via Optical-Flow Diffusion}

\author{Manuel Serna-Aguilera\\
    University of Arkansas\\
    1 University of Arkansas\\
    {\tt\small mserna@uark.edu}
    \and
    Second Author\\
    Institution2\\
    First line of institution2 address\\
    {\tt\small secondauthor@i2.org}
}
\maketitle

%%%%%%%%% ABSTRACT
%\begin{abstract}
%   xxx
%\end{abstract}

%%%%%%%%% BODY TEXT
%--------------------------------------------------
\section{Cell Annotation Program}
The annotation process to quantitatively test the area, perimeter, and neurite length distributions was performed via a published MATLAB program (see References); Figure \ref{fig:annotation_matlab} gives a visualization. The cell body centers are first selected (numbered red star points), and then the cell body is traced (yellow patches with fill). The long axis of each neuron is chosen by the user and plotted with a white line. The cell body will be reoriented so that this white line points in the ``up'' direction for shape comparison, shown in Figure 6 in the paper. The perimeter for each cell body is calculated by adding the lengths between each point, and the area is calculated using the \texttt{polyarea()} built-in function. Neurites can then be traced from the cell body, and neurite branches can also be added from existing neurites. Neurites are in magenta if self-terminated or cyan if connected to other neurons, and additional branches are traced in red. The neurite diameter at each point is estimated and visualized using green circles based on a 4\% contrast difference cutoff between the selected neurite point and its neighboring pixels. Green points plotted on each circle indicate the precise pixel that triggered the 4\% contrast difference cutoff. Neurites are labeled at the point of termination with the neuron number (N, followed by a number), neurite number on that neuron (n, followed by a number), and branch number (B, followed by a number). A conversion ratio between pixels and microns of 1.1939, measured using a scale bar in the raw images, is used post-calculation to convert to micron measurements. Neurites' lengths and directions that connect between two neurons are counted twice (twice because neurites come from both cells and join), and directions are normalized against the shortest segments in each trace map to help ensure comparable histogram plots.

%--------------------------------------------------
\section{Cell Area and Perimeter Distributions}
Figure \ref{fig:dist_comparison} shows violin plots of the raw, training, and synthesized cell body areas for both neuron cells and dead cells as calculated by the MATLAB-based annotation program. A black boxplot in the center of the violin indicates interquartile range. The median and 95\% confidence intervals of the median for each dataset are indicated with white circles and triangles, respectively. The p-values, as calculated with the two-sample t-test, are shown between each dataset. Values falling above $p = 0.05$ are labeled as ``n.s.'' for ``not significant''. As shown, the distribution shapes and medians for each dataset are comparable. Within each group, the differences between the training and synthesized datasets are not statistically significant (neuron perimeter is slightly significant with $p = 0.044$, and n.s. for all other cases). This suggests the model does a fair job synthesizing comparable datasets from the training data. The differences between the raw and synthesized datasets for neuron area, dead cell area, and dead cell perimeter are not statistically different, but the neuron perimeter is statistically different ($p = 0.014$). Therefore, the synthesized data does not match the raw data as closely and can be improved further. The differences between the raw and training datasets for neuron perimeter, dead cell perimeter, and dead cell area are not statistically different ($p =$ n.s.), but the neuron area is statistically different ($p = 0.005$). The training data is therefore not representative of the raw dataset. This is most likely due to cropping of large cell bodies in the training dataset. The training dataset will need to be improved to ensure models that correctly synthesize neuronal behavior.

%--------------------------------------------------
% Fig: neuron MATLAB annotation visualization
\begin{figure*}%[t]
    \centering
    \includegraphics[width=1.0\linewidth]{images/supplementary/MyExperiment_WHITE_001686 Tracemap_example1.png}
    \caption{
        MATLAB-based annotation of neural stem cells from the phase contrast images. Enlarged to show details. Best viewed in color.
    }
\label{fig:annotation_matlab}
\end{figure*}

% Fig: Area and Perimeter distributions
\begin{figure*}
  \centering
  %\begin{subfigure}{0.9\linewidth}
  \begin{subfigure}{1.0\linewidth}
    \includegraphics[width=1.0\linewidth]{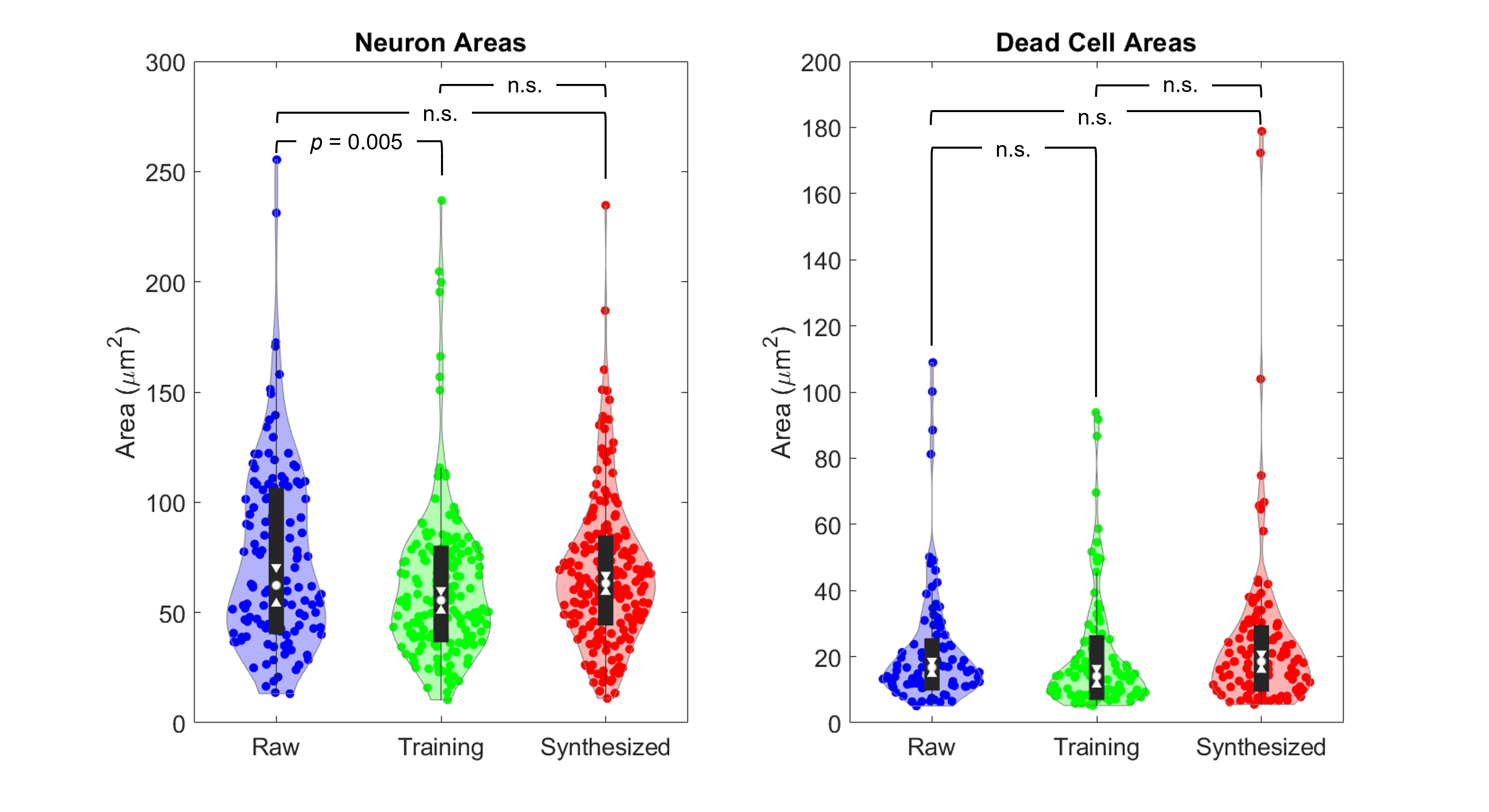}
    \caption{Area distributions for cells in the raw videos (blue), training patch videos (green), and synthesized videos (red).}
    \label{fig:area_dist}
  \end{subfigure}
  \hfill
  %\begin{subfigure}{0.9\linewidth}
  \begin{subfigure}{1.0\linewidth}
    \includegraphics[width=1.0\linewidth]{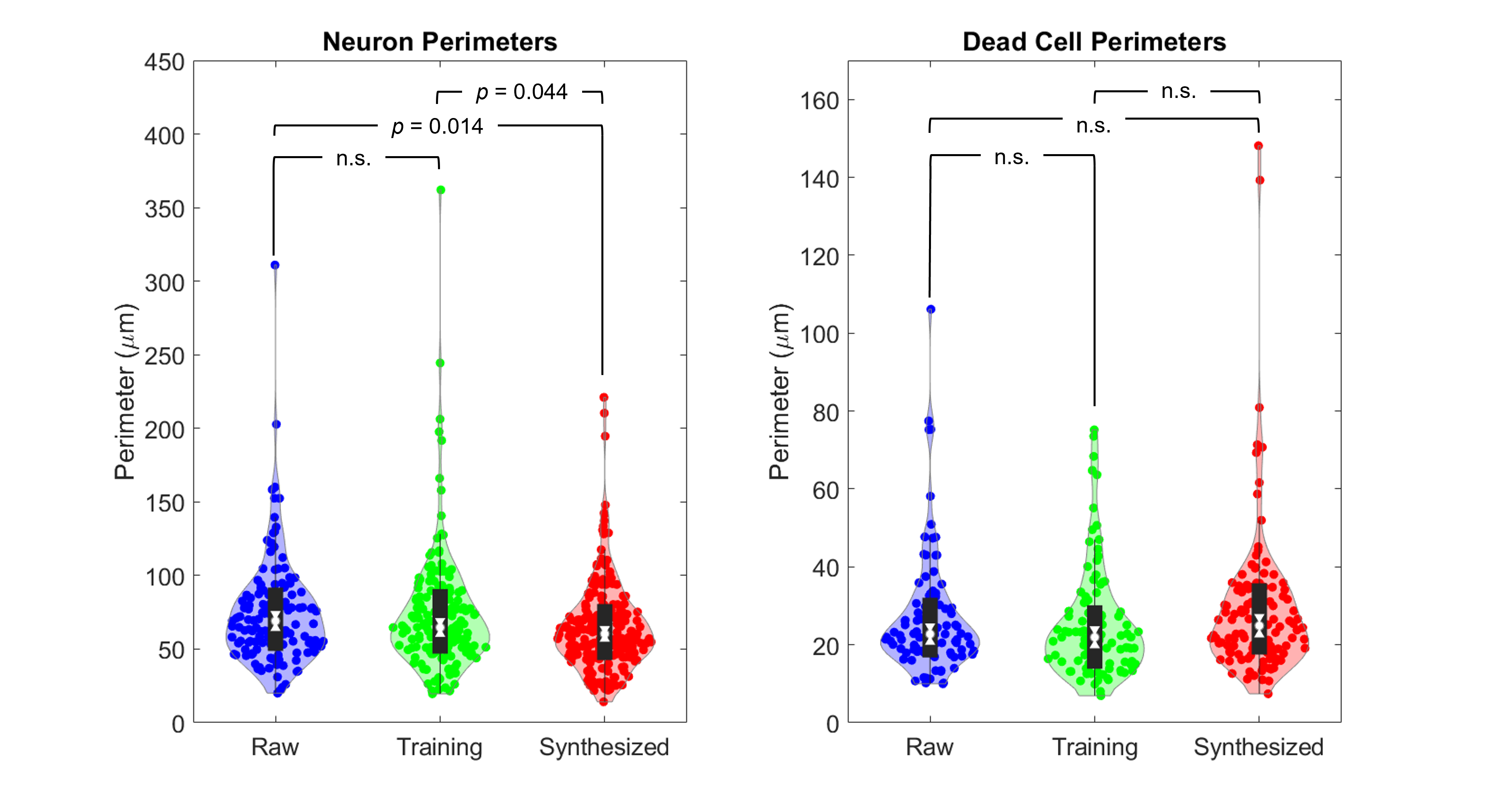}
    \caption{Perimeter distributions for cells in the raw videos (blue), training patch videos (green), and synthesized videos (red).}
    \label{fig:perimeter_dist}
  \end{subfigure}
  \caption{Neuron and dead cell body areas in \ref{fig:area_dist} (in square micrometers, $\mu m^2$) and perimeters in \ref{fig:perimeter_dist} (in micrometers, $\mu m$), as calculated by the annotation program. Note ``n.s.'' denotes ``not significant''.
  }
  \label{fig:dist_comparison}
\end{figure*}
%--------------------------------------------------

%%%%%%%%% REFERENCES
% NOTE: don't think I need refs
%{\small
%\bibliographystyle{ieee_fullname}
%\bibliography{egbib}
%}